\documentclass[letterpaper, 10 pt, conference]{ieeeconf}  
\IEEEoverridecommandlockouts 
\usepackage[utf8]{inputenc}
\usepackage{cite, epsfig, mathptmx, times, amssymb, amsmath, color}
\usepackage{graphicx,booktabs,soul,color,graphics,epsfig,mathptmx,moreverb,ifpdf,cancel,framed,dblfloatfix}
\usepackage{gensymb,graphicx,url, multirow}
\usepackage{pbox}
\usepackage{adjustbox}
\usepackage[linesnumbered,ruled,vlined]{algorithm2e}
\usepackage[x11names,dvipsnames,table]{xcolor}
\usepackage{colortbl}
\usepackage{subcaption}
\usepackage{floatrow}
\usepackage{float}
\usepackage{balance}
\usepackage{xcolor}
\usepackage{amsmath,amssymb,amsfonts}
\usepackage{textcomp}
\usepackage{lipsum}
\usepackage{siunitx}

\DeclareMathAlphabet{\mathpzc}{OT1}{pzc}{m}{it}

\title{\LARGE \bf MRI-powered Magnetic Miniature Capsule Robot \\with HIFU-controlled On-demand Drug Delivery }

\author{Mehmet Efe Tiryaki$^{1,2}$, Fatih Doğangün$^{1}$, Cem Balda Dayan$^{1}$, Paul Wrede$^{1,3}$, Metin Sitti$^{1,2,4}$
	\thanks{This work is funded by the Max Planck Society.}
	\thanks{$^{1}$ Physical Intelligence Department, Max Planck Institute for Intelligent Systems, 70569 Stuttgart, Germany. E-mail: sitti@is.mpg.de}
	\thanks{$^{2}$ Institute for Biomedical Engineering, ETH Zurich, 8092 Zurich, Switzerland}
	\thanks{$^{3}$ Max Planck and ETH Center for Learning Systems, Stuttgart, Germany }
	\thanks{$^{4}$ College of Engineering and School of Medicine, Koç University, 34450 Istanbul, Turkey.}
}

\begin{document}

\maketitle

\begin{abstract}

Magnetic resonance imaging (MRI)-guided robotic systems offer great potential for new minimally invasive medical tools, including MRI-powered miniature robots. By re-purposing the imaging hardware of an MRI scanner, the magnetic miniature robot could be navigated into the remote part of the patient's body without needing tethered endoscopic tools. However, the state-of-art MRI-powered magnetic miniature robots have limited functionality besides navigation. Here, we propose an MRI-powered magnetic miniature capsule robot benefiting from acoustic streaming forces generated by MRI-guided high-intensity focus ultrasound (HIFU) for controlled drug release. Our design comprises a polymer capsule shell with a submillimeter-diameter drug-release hole that captures an air bubble functioning as a stopper. We use the HIFU pulse to initiate drug release by removing the air bubble once the capsule robot reaches the target location. By controlling acoustic pressure, we also regulate the drug release rate for multiple location targeting during navigation. We demonstrated that the proposed magnetic capsule robot could travel at high speed up to 1.13 cm/s in ex vivo porcine small intestine and release drug to multiple target sites in a single operation, using a combination of MRI-powered actuation and HIFU-controlled release. The proposed MRI-guided microrobotic drug release system will greatly impact minimally invasive medical procedures by allowing on-demand targeted drug delivery.

\end{abstract}


\section{Introduction}

\begin{figure}[t]
	
	\vspace{1.8 mm}
	\includegraphics[width=\linewidth]{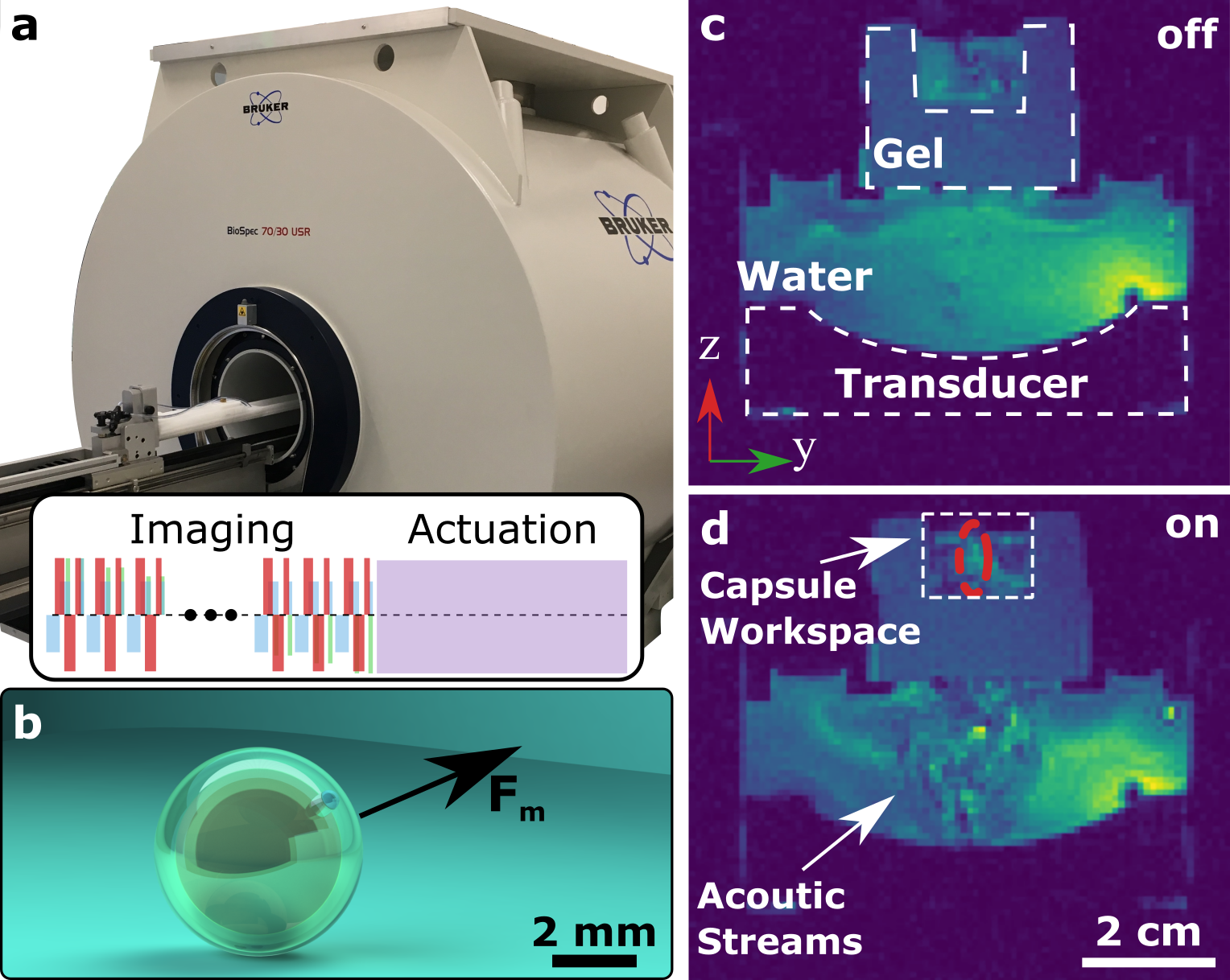}
	\caption{MRI-powered magnetic miniature capsule robot with high-intensity focused ultrasound (HIFU)-controlled drug release. a) 7 Tesla small animal MRI scanner with 2D imaging and actuation MRI sequence. The pulse sequence is composed of imaging and actuation gradients. b) Miniature magnetic capsule robot design with air bubble captured in the hole. The black vector demonstrates an arbitrary 3D magnetic actuation force applied by the MRI scanner. c,d) The axial MR images of HIFU during it is off (c) and on (d). The on-phase of the HIFU can be seen by motion of the water in MR images.\vspace{-0.6cm}}

\end{figure}
Recently, MRI scanners have emerged as a robotic platform for magnetic miniature robots, combining high-quality medical imaging and magnetic actuation in a single medical device \cite{erinreview}. Researchers re-purposed the imaging gradient coil of the MRI scanner to generate magnetic pulling force enough to navigate magnetic miniature robots in body fluids \cite{tiryaki2021magnetic, tiryaki2022deep, martel2007automatic, folio2017two}. MRI-powered magnetic miniature robots demonstrated great potential for minimally invasive medical operations, such as MRI-guided microrobotic drug delivery. However, the functionality of these magnetic miniature robots has been hindered by MRI scanners' extremely constrained magnetic and RF conditions limiting additional remote stimuli mechanisms.

High-intensity focused ultrasound (HIFU) gained attention as an MRI-compatible remote stimuli mechanism for controlled drug delivery\cite{thanou2013mri}. Focused acoustic waves could trigger drug release by thermal, hydrodynamic, and cavitation effects allowing vast design possibilities for drug delivery systems\cite{yildirim2019colloids,wei2021ultrasound,matoori2019mri,grull2012hyperthermia,zhang2014continuous,zhou2014ultrasound,jeong2020acoustic} and also can enhance drug diffusion by opening physiological barriers, such as blood-brain barrier \cite{fan2013spio}. In addition, integrating HIFU with MRI-guided robotic systems enabled researchers to perform accurate drug release triggers in the soft tissue with minimal damage to surrounding tissue\cite{dai2021robotic}. However, MRI-guided HIFU-based drug delivery systems have been limited to passive targeting mechanisms, and the potential of MRI-powered magnetic microrobotic navigation has not been demonstrated.

In this paper, we brought together the navigation capabilities of MRI-powered magnetic microrobotic systems and the acoustic triggering of the MRI-guided HIFU system. First, we turned the MRI scanner into a magnetic microrobot actuation system by modifying a commercial MR imaging sequence with an additional externally controlled magnetic gradient signal (Fig. 1a). Then, we designed a drug-carrying polymeric capsule robot with a single hole (Fig. 1b). The hole serves two purposes in our design; first, capturing an air bubble functioning as a stopper due to the hydrophobic surface property; second, releasing the drug with the help of acoustic streaming forces. The drug release could be initiated by removing the air bubble using targeted HIFU pulses and further regulated by controlling acoustic pressure. We characterized the effect of hole geometry on required acoustic pressure and the drug release rate as a function of acoustic pressure. We built an MRI-compatible 2D piezo stage with a stationary HIFU transducer that could be operated during MR imaging and actuation (Fig. 1c-d). Finally, we demonstrated MRI-guided navigation of the miniature capsule robot in the porcine small intestine and drug release to multiple target locations during MRI-powered navigation in an in vitro environment.

\subsection{ Related Research}

One of the main challenges in microrobotic drug delivery systems is combining a functional drug-releasing mechanism with a microrobotic platform capable of locomotion, and localization \cite{sitti2015biomedical}. MRI scanner as a microrobotic platform is one of the promising candidates for overcoming this challenge by providing magnetic actuation and high-resolution medical images in a single device. However, integrating any drug delivery method to MRI-powered miniature robots requires mutual compatibility of the release mechanism, MR imaging, and magnetic actuation.

On-demand drug delivery with magnetic miniature robots has been achieved either by magnetic force-induced deformation \cite{yim2011design,zhang2021voxelated,le2016soft} or external stimuli, such as radiofrequency (RF) field \cite{hu2011remotely,pradhan2010targeted,koo2020wirelessly}, light \cite{bozuyuk2018light,alapan2020multifunctional}, or acoustic pulses \cite{matoori2019mri,grull2012hyperthermia,zhang2014continuous,zhou2014ultrasound,wei2021ultrasound,jeong2020acoustic}. Magnetic force-induced drug release capsules either mechanically pump the drug out \cite{yim2011design,zhang2021voxelated} or open the drug chamber to release the drug at once \cite{le2016soft}. Due to the relatively large size, these capsules can travel large distances in body cavities, such as the gastrointestinal (GI) tract, and can deliver a higher volume of drugs. The magnetic force-induced pumping mechanism can even perform drug release to multiple locations. However, magnetic force-induced drug release systems couple magnetic actuation with drug release triggering. Especially, deformation-based pumping mechanisms requires high magnetic field gradient with independently controlled magnetic field direction, which is not possible to achieve with MRI scanners' coil systems.

The alternative is decoupling drug release from MRI-based magnetic actuation using external stimuli. RF field is a commonly used stimulus that does not affect magnetic actuation when applied at high frequencies \cite{hu2011remotely,pradhan2010targeted,koo2020wirelessly}. The RF field could be used for heating-based drug release by direct magnetic and joules heating \cite{hu2011remotely,pradhan2010targeted}, or wireless power transfer \cite{koo2020wirelessly}. Direct heating is generally used in nano-scaled systems with temperature-sensitive organic capsules like liposomes. While such nano-scaled systems can theoretically pass through any microchannels in the human body, the magnetic gradient forces in MRI scanners are not strong enough to actuate such small systems in practice. On the other hand, wireless power transfer systems could be integrated with larger magnetic robotics systems allowing easy actuation. However, introducing too much magnetic material in robot design obstruct MR imaging due to large image artifacts. Besides magnetic actuation considerations, RF field-based drug release systems require high-power external RF circuits. Since such RF circuits interfere with MR imaging and completely obstruct MR images, they are not preferred in MRI scanners. Light-triggered drug release is another alternative for magnetic microrobotic systems, which do not interact with either MR imaging or actuation \cite{bozuyuk2018light,alapan2020multifunctional}. However, light-triggered drug release systems suffer limited penetration depth; hence, they cannot be used in deep tissue.

HIFU-based acoustic triggering received the most attention in MRI-guided drug release applications due to its high penetration depth and MRI compatibility in terms of magnetic and RF interactions\cite{wei2021ultrasound}. Focused acoustic waves could induce drug release through many mechanisms depending on the desired size scale. At the nano/microscale, many studies proposed cavitation-based \cite{zhou2014ultrasound} and thermoresponsive \cite{grull2012hyperthermia,zhang2014continuous} systems to mediate drug release demonstrating great potential for targeted delivery. However, these nano/microscale systems suffer insufficient actuation forces for navigation as other nanosystems.

At the larger scale, thermal phase change-based \cite{matoori2019mri} and cavitation-based pumping \cite{jeong2020acoustic} have been used for triggering. Due to the larger size, these systems are easy to track with MR imaging and magnetic actuation. For instance, Matoori et al. have demonstrated MRI-guided remote triggering with HIFU pulse-induced phase change\cite{matoori2019mri}. However, \cite{matoori2019mri} has only demonstrated the passive capsule motion. On the other hand, Jeong et al. introduced magnetic components to their system to navigate the capsule to the target location without integrating any medical imaging, such as MR imaging\cite{jeong2020acoustic}. To our knowledge, there has not been any MRI-based microrobotic drug delivery system that brings together MRI-guided HIFU-based drug release and MRI-powered magnetic actuation for targeted drug delivery.

\subsection{Contribution }

Here, we present a proof-of-concept MRI-powered magnetic miniature capsule robot with HIFU-controlled drug release. We proposed an easy-to-manufacture magnetic capsule robot design benefiting acoustic streaming forces to trigger drug release and regulate release rate. Our main contributions are:
\begin{itemize}
	\item MRI-powered actuation with MRI-compatible HIFU-based drug delivery;
	\item air bubble-based HIFU-triggered drug release activation mechanism;
	\item HIFU-controlled drug release for multiple location targeting.
\end{itemize}

\begin{figure}[t]
	\vspace{1.8 mm}
	\includegraphics[width=\linewidth]{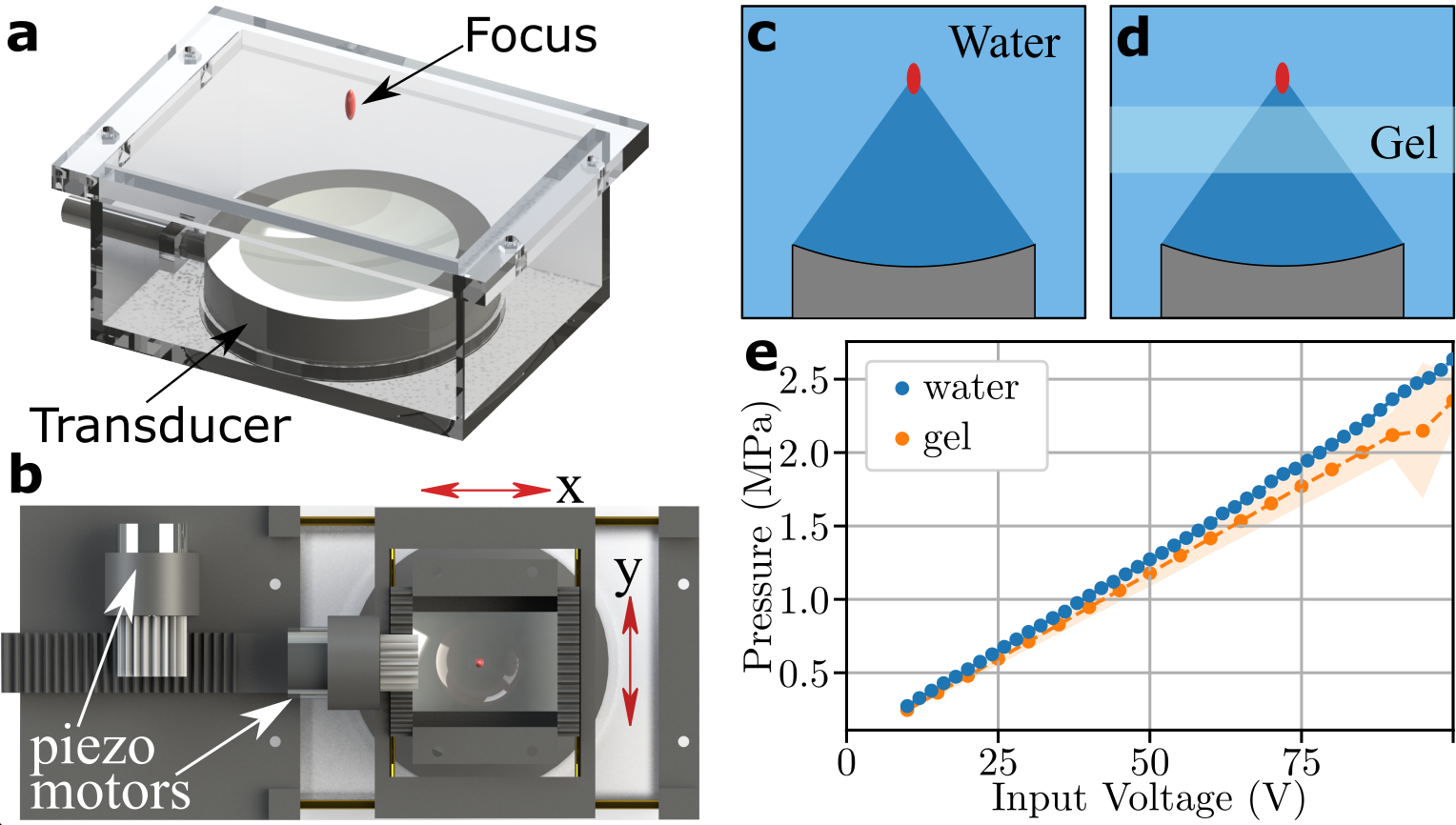}
	\caption{2D motorized HIFU system and pressure characterization. a) The HIFU transducer fixed in water bath. The red ellipsoid is the focal point of the HIFU tranducer. b) Top view of 2D motorized piezo stage placed on the water bath with HIFU transducer. The schematic illustration of HIFU focus c) through only water and d) through water and 1.5 cm thick $2\%$ agarose gel. e) The acoustic pressure measurement at focal point.\vspace{-0.6cm}}
\end{figure}

\section{MRI-powered Magnetic Actuation}

MRI scanner comprises three sets of electromagnetic coils that contribute to MR imaging. First, the main super-conducting magnet sets a highly uniform magnetic field at the center of the MRI scanner to spin hydrogen atoms at their Larmour frequency. Next, the radiofrequency (RF) coil excites the spins to generate the signal. Finally, the gradient coils encode the spatial information to obtain MR images. Once a magnetic object enters the uniform magnetic field, it causes a large artifact around its vicinity, which allow us to spot the object in the MR image easily. The size of the artifact is based on imaging parameters and the amount of the magnetic core \cite{tiryaki2021magnetic}. The robot is located roughly at the center of the artifact, except during motion \cite{tiryaki2022deep}.

Besides imaging, the MRI scanner's magnetic gradient coils could be repurposed for magnetic actuation by modifying the MRI pulse sequence in an alternating imaging and actuation scheme \cite{tiryaki2021magnetic}. The high field of the MRI main coil magnetically saturates and aligns the magnetic core of the robot with the high magnetic field. Therefore, the magnetic capsule robot can be actuated by magnetic gradient pulling force in 3D space. The force exerted on the robot can be calculated as:

\begin{equation}
\mathbf{F}_m = m_s \mathbf{G},
\end{equation}

\noindent where $m_s$ is the magnetic moment at saturation magnetization, and  $\mathbf{G}=\nabla(\hat{x}\cdot \mathbf{B})$ is the magnetic field gradient in $\hat{x}$ direction, i.e. axial direction of the MRI scanner.

In this study, we performed all our experiments in a 7T small animal MRI scanner (BioSpec
70/30, Bruker). We used a modified gradient-recalled echo (GRE) sequence by adding a real-time adjustable gradient pulse after every 2D MR imaging pulse group, as shown in Fig. 1a. Since the gradient coils also apply magnetic pulling force during the imaging part of the MRI sequence, we modified the imaging pulses to create zero net magnetic force at the end of each imaging period \cite{tiryaki2022deep}. The sequence parameters used in experiments were: pixel number 128, slice thickness \SI{2}{\milli\meter}, flip angle \SI{40}{\degree}, RF pulse bandwidth \SI{10}{\kilo\hertz}, receiver bandwidth \SI{400}{\kilo\hertz}, 2D image duration \SI{1000}{\milli\second}, echo time \SI{1.61}{\milli\second}, repetition time \SI{7.81}{\milli\second} , imaging time \SI{680}{\milli\second}, actuation time \SI{320}{\milli\second}. We also limited the maximum actuation gradient \SI{60}{\milli\tesla/\meter} to operate within the limits of magnetic gradients available in clinical MRI scanners.

\begin{figure}[b]
	\includegraphics[width=\linewidth]{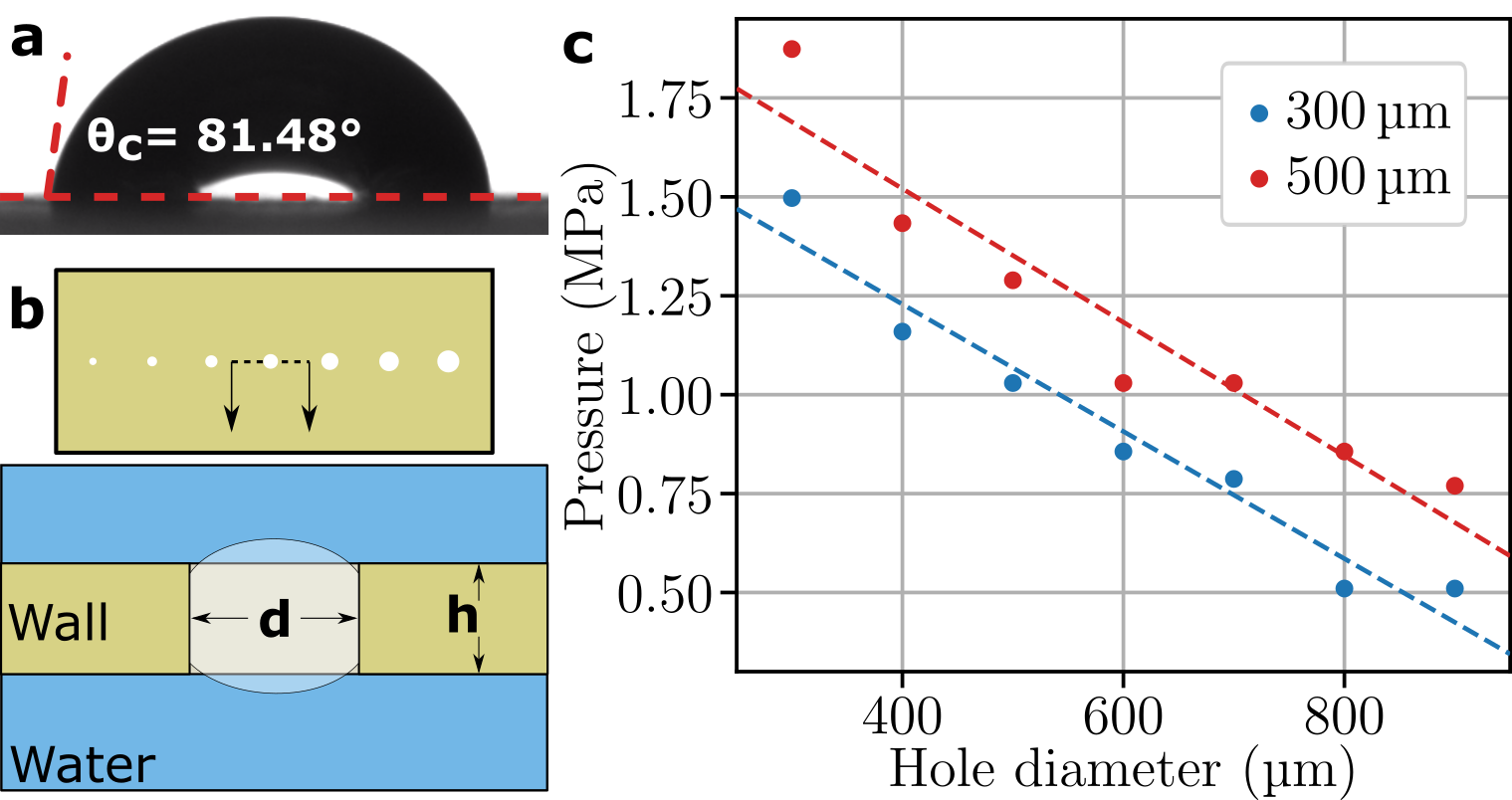}
	\caption{Bubble formation and removal characterization. a) DI-water droplet on the flat substrate, which is made of the same material with the 3D printed polymeric robot shell. b) The bubble-removal experiment setup. Upper figure is the flat plates with multiple holes with different diameters. Lower figure is the cross-section of hole shown in upper figure with air bubble. The bubble has a convex meniscus due the hydrophobic surface. $d$ is the hole diameter and $h$ is the wall thickness c) Experimental measured required acoustic pressure as function of hole geometry. The dash lines are linear fitting lines.}
\end{figure}

\section{MRI-compatible HIFU}
We used an MRI-compatible constant curvature HIFU transducer with 5 cm focal distance (H-104, Sonic Concepts). Although high degree of freedom in MRI-compatible robotic HIFU systems exists \cite{dai2021robotic}, due to the limited space in our small animal MRI scanner, we fixed our HIFU transducer at the bottom of a rectangular water pool (Fig. 2a). Then, we built a 2D MRI-compatible piezo motion stage to move our sample in the $x$-$y$ plane relative to the HIFU transducer (Fig. 2b). We used a function generator (AFG1022, Tektronix) and power amplifier (Model 150100C, Amplifier Research, Germany)  with 100 gain to generate a 500 kHz sinusoidal input signal up to 100 V. We tested the MRI-compatible HIFU setup with a 2D stage in MRI (Fig. 1c) and observed the location of the focus in real-time MR images (Fig. 1d).

Later, we characterized the acoustic pressure at the focus position using a hydrophone in water-only (Fig. 2c) and through-gel (Fig. 2d) conditions. We first placed the hydrophone at the focal point by maximizing the echo signal read from the hydrophone. Then, the input voltage was swept between 10 to 100 V with 2 V increments, resulting in a linear trend (Fig. 2e). Next, we repeated the same measurement through a 1.5 cm thick $2\%$ agarose gel (Type I-A, Sigma-Aldrich). In order to average the effect of inhomogeneities in the gel, we measured pressure at 20 different points on the gel. The mean and standard deviation is shown in Fig. 2e, demonstrating a slight pressure drop due to the absorption of the gel. We used achieved characterization curves to estimate the acoustic pressure in the rest of the paper.

\section{Magnetic Miniature Capsule Robot}

To understand the air bubble formation on the hole,  we measured the static contact angle using the sessile drop method by commercial goniometer (Drop Shape Analyzer DSA100, Krüss GmbH, Hamburg, Germany). We printed flat substrates with the material of the capsule shell (Clear V4, Formlabs), using an Stereolithography  (SLA) 3D printer (Form3, Formlabs). During these measurements, deionized water (DI-water) was injected with a controlled volume ($\approx$\SI{2}{\micro\liter}) at a constant rate on the flat substrate. After dosing, the DI-water droplet was kept on the flat substrate for around 30 seconds, and then the contact angle was measured. The static contact angle was measured 10 times at different locations on the substrate. The contact angle of DI-water on the 3D printed flat substrate is 81.48$\pm$ 1.71 \SI{}{\degree} contact angle, showing a near hydrophobic surface characteristic (Fig. 3a). Then, we printed flat plates with \SI{300}{}-\SI{500}{\micro\meter} thickness and multiple holes with diameter between \SI{300}{} to \SI{1000}{\micro\meter}, as shown in Fig. 3b. Once the plates were submerged in water, we observed that a bubble was formed in all but \SI{1000}{\micro\meter}-diameter hole with a concave meniscus (similar to Fig. 3b), which was expected due to the surface hydrophobicity and capillary effect. Later, we placed each hole at the focal point of the HIFU. We ramped the acoustic pressure on the bubble with 200 kPa increments starting from 500 kPa and recorded the pressure at which the bubble was removed. We repeated experiments on 2 different plates for each thickness by reforming the bubble 5 times for each hole diameter. We observed nearly linear relation between hole diameter and pressure (Fig. 3c), where required acoustic pressure decrease with increasing hole size. We also saw that the increasing wall thickness increases the required acoustic pressure.

We finalized our design with \SI{5}{\milli\meter} capsule diameter, and \SI{500}{\micro\meter} by \SI{500}{\micro\meter} wall thickness and hole diameter. The manufacturing process of the capsules is described in Fig. 4a. We first started by 3D printing two hemispherical shells, one with a circular hole (Fig. 4a.i), using the SLA printer. We verified the size of the hole with a needle. Then we assembled two hemispherical shells with a \SI{1.5}{mm}-diameter stainless steel magnetic core (AISI420, Kugel-Winnie, Bamberg, Germany)  at the center by applying UV-cured resin (Fig. 4a.ii). Due to the capillary surface force of resin, we did not need an additional alignment for the hemispherical shells. Later, we placed the assembled capsule hole facing upward under UV for 45 minutes (Fig. 4a.iii). The cured capsule was manually filled with food dye (Bakeryteam, Germany) to represent a water-soluble drug with a needle (Fig. 4a.iv). The \SI{300}{\micro\meter}-diameter needle was chosen to be smaller than the hole to allow air to leave the cavity. The excess drug droplet leaking from the hole is cleaned with a tissue to keep the drug-air interface at the bottom of the hole. Then the drug-filled capsule was air-dropped into the water to form the stopper bubble in the hole (Fig. 4b.i). It is important to note that the tissue cleaning process is necessary for the repeatability of the air bubble capturing process since it allows air to remain in the hole after filling. Moreover, a dry robot surface is essential for final meniscus creation \cite{dayan20213d} (Fig. 4b.ii). It is not possible to capture air bubbles when the capsule surface is fully wetted.

After manufacturing, we tested the robustness of the air-bubble stopper using a magnetic steerer at 500 rpm, showing that translational or rotations of the capsule do not remove the air bubble. We also checked long term storage the drug-loaded capsule. We placed 4 drug-loaded capsules with and without an air bubble in DI water and refrigerated them at \SI{4}{\celsius}. While the capsules without air bubbles leaked all drugs in 1 day, we did not observe any leakage from capsules with air bubbles after 4 days.

Later, we tested our drug-loaded capsule robot for HIFU triggering. We first investigated the required acoustic pressure to remove the bubble by placing the robot on an agarose gel pool. The air bubble was successfully removed at 1.25 MPa acoustic pressure (Fig. 4b.iii). Then, we characterized the drug release rates of 5 different capsule robots. We first filled the capsules with dye, and the capsule robot was pulled towards the wall of the agarose pool with the help of an external magnet. Then we measured the time required for fully emptying the capsules under 3 different acoustic pressure. The average emptying times are given in Table I. Then we calculated the release rate by dividing cargo volume by the release time (Fig. 4c). Although higher acoustic pressures could release all drugs within less than 1 minute, we observed that the acoustic radiation forces push the capsule away from the HIFU focus significantly. Therefore, the operator needed to adjust the capsule position continuously. On the other hand, for lower pressure (\SI{1.4}{\mega\pascal}), the drug was released without moving the capsule significantly.

\begin{figure}[t]
	
	\vspace{1.8 mm}
	\includegraphics[width=\linewidth]{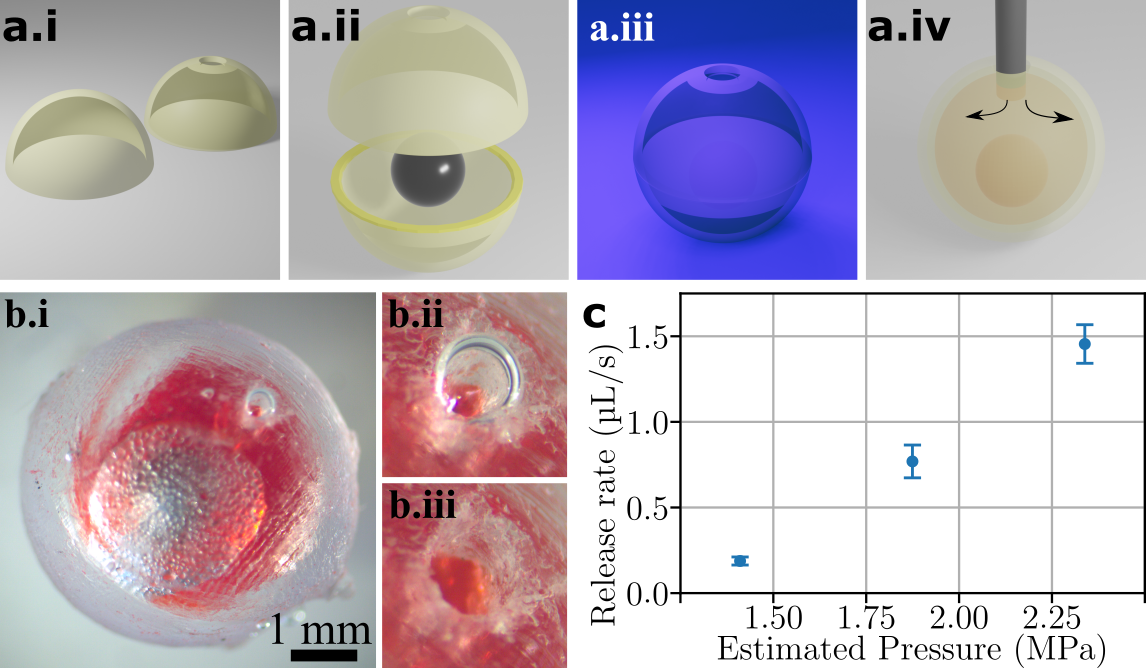}
	\caption{Capsule manufacturing and characterization. a) Manufacturing schematic of the drug delivery capsule. a.i) 3D printed hemispherical shells. One hemisphere has the hole for drug release. a.ii) Assembly with magnetic core. The magnetic core is placed in the shells and UV-cured resin is applied.  a.iii) UV curing.  a.iv) Filling the drug. The drug is filled by a needle slightly smaller than the hole. b) The manufactured \SI{5}{\milli\meter}-diameter spherical capsule robot with \SI{1.5}{\milli\meter} diameter magnetic core. The capsule wall thickness is \SI{500}{\micro\meter} and the hole diameter \SI{500}{\micro\meter}. b.i) Submerged capsule with the bubble. Close-up view of the bubble b.ii) before HIFU-triggered air bubble removal, b.iii) after HIFU-triggering. c) The controlled release rate of capsule robot with respect to estimated acoustic pressure. \vspace{-0.7cm}}
\end{figure}
 
\begin{table}[b]
	\vspace{-0.3cm}
	\centering
	\small
	\tabcolsep=0.1cm
	\caption{Capsule emptying duration}
	\footnotesize
	\begin{tabular*}{\textwidth}{p{0.31\textwidth}p{0.2\textwidth}p{0.2\textwidth}p{0.2\textwidth}}
		\rowcolor[rgb]{ .867,  .922,  .969}
		Pressure (MPa) & 1.4 & 1.9 & 2.4 \\
		Emptying Duration (s) &  $158\pm18$ &  $38\pm5$ & $20\pm2$ \\
	\end{tabular*}
\end{table}

\section{Experiments and Results}
\subsection{MRI-powered navigation}

To demonstrate the capsule robot's remote navigation capability, we performed open-looped MRI-powered magnetic actuation experiments in the \textit{ex vivo} porcine small intestine. First, we filled 50 cm long porcine small intestine with Ringer solution (Deltamedica, Reutlingen, Germany) and placed it in a box in spiral form, as seen in Fig. 5. We inserted the capsule robot in the small intestine. Then, we started MRI imaging and actuation sequence, and the human operator navigated the capsule from one end to the other using a joystick and visual feedback from real-time MR images. We repeated the navigation experiment 5 times and recorded $44.2\pm10.1$ s average navigation time, resulting in an average speed of \SI{1.13}{\centi\meter/\second}. We observed that most of the time lost during navigation occurred around the turning points, where the small intestine folds on itself.

\subsection{Multiple target drug release}

Later, we demonstrated the HIFU-controlled drug release with MRI-powered actuation. We placed the drug-loaded capsule robot without an air bubble in an U-shaped agarose gel channel (Fig. 6). Since the magnetic distortion of the robot obstructs drug release in the MR images, we used an MRI-conditioned camera (MRC systems, Heidelberg, Germany) to verify drug release visually. The capsule robot was navigated to 4 different target positions using MRI-powered magnetic control. Then, the HIFU focus point was aligned with the capsule robot using the 2D piezo stage, and an acoustic pulse with 1.4 MPa pressure was applied to the capsule. The drug was released successfully to all four target points. We also observed that the drug was perfused by the acoustic pulses, as seen in Fig. 6. During the experiment, we see that the acoustic radiation forces mildly pushed the capsule robot away from the target site. The operator needed magnetic forces to keep the robot at the desired position.

\begin{figure}[t]
	
	\vspace{2 mm}
	\includegraphics[width=\linewidth]{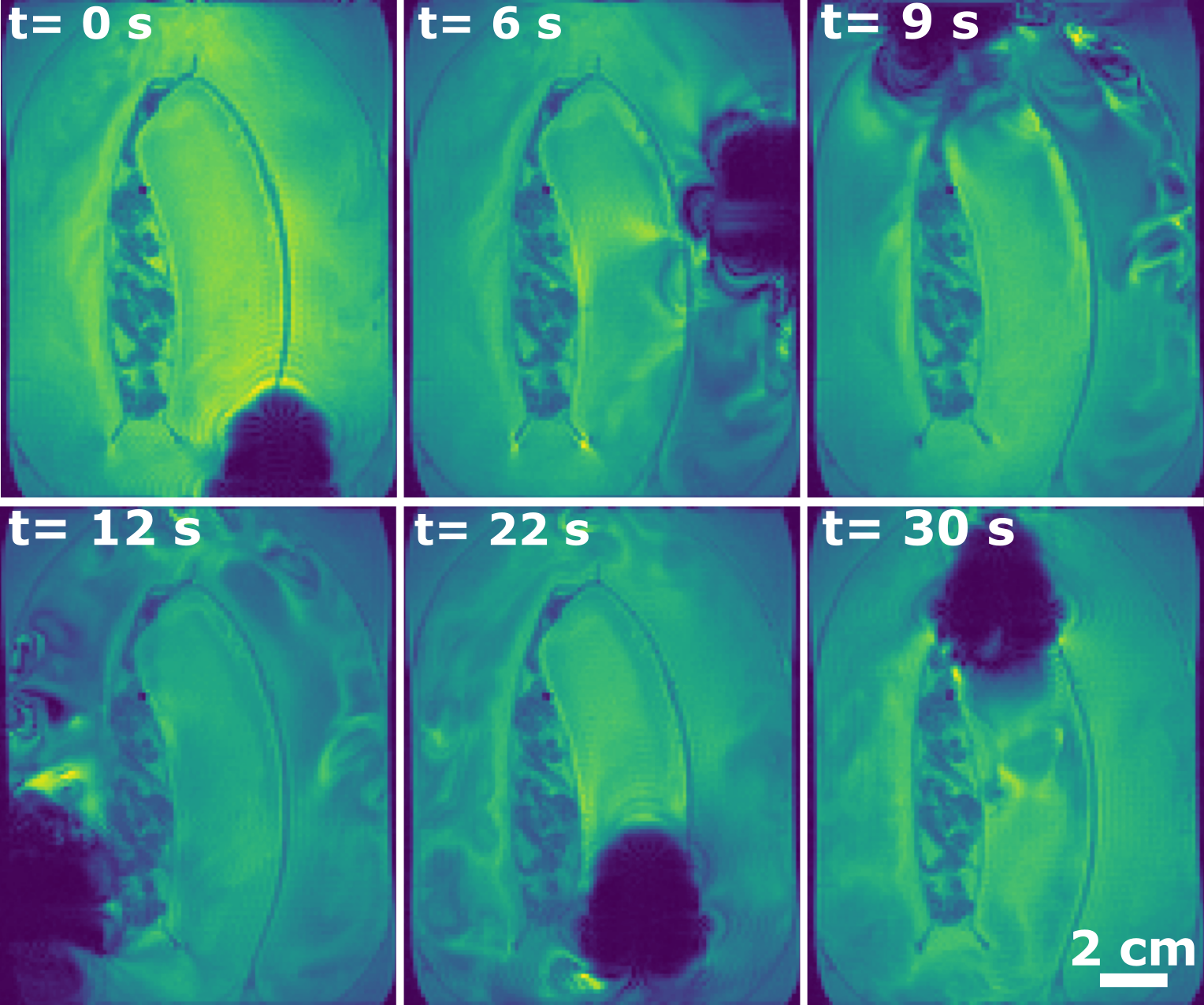}
	\caption{MR image snapshots of real-time MRI-powered navigation in the porcine small intestine. The magnetic capsule robot is actuated using a modified GRE sequence with magnetic actuation. The black artifact is the signature of the 1.5 mm magnetic core in the MR image.\vspace{-0.7cm}}
\end{figure}

\section{Conclusion}
This paper demonstrates an MRI-guided magnetic microrobotic system bringing MRI-powered magnetic actuation and MRI-guided HIFU-controlled drug release. We showed that MRI-powered magnetic miniature capsule robots \cite{tiryaki2021magnetic,tiryaki2022deep} can be functionalized for drug release using  MRI-guided HIFU. We demonstrated a proof-of-concept air bubble-based drug stopper mechanism and HIFU-controlled drug release concept. In our characterizations, we observed that the drug-release hole geometry directly affects the required acoustic pressure. Decreasing hole size increases the required pressure while decreasing wall thickness decreases the pressure. Later, we also characterized the HIFU-controlled drug release rate as a function of acoustic pressure using a magnetically stabilized capsule, showing a nearly linear relation between acoustic pressure and drug release rate.

Furthermore, we demonstrated MRI-powered navigation of the magnetic capsule robot in \textit{ex vivo} porcine small intestine with relatively high speeds of \SI{1.13}{\centi\meter/\second}. Achieved speed was approximately two orders of magnitude faster compared to the average travel speed of passive capsule robot, which is \SI{0.025}{\centi\meter/\second} reported by \cite{worsoe2011gastric}. Such high speeds could allow surgeons to transport drug-loaded capsules on-demand to any location in the GI tract in a couple of minutes rather than waiting a couple of hours for passive approaches. Moreover, the demonstrated HIFU-controlled drug release mechanism allows multiple localized micro drug release operations in a single operation. Such localized micro drug release could be beneficial in targeting multiple malignant cysts in the small intestine at the early stage.

\begin{figure}[b]
	\includegraphics[width=\linewidth]{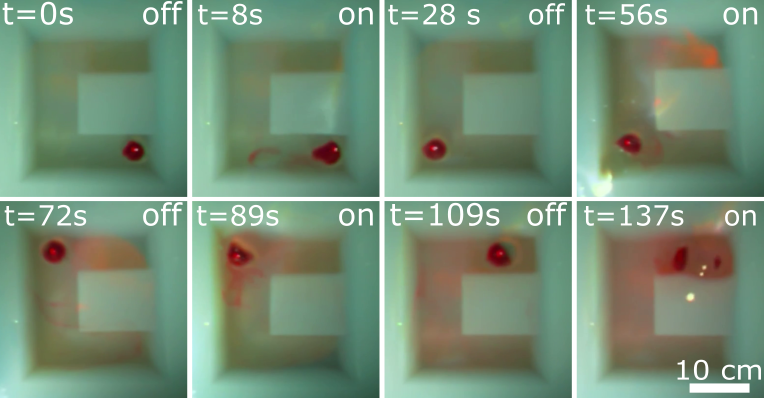}
	\caption{Snapshots of HIFU-controlled drug release at multiple targets using MRI-powered actuation. The on/off state of the HIFU is shown on each snapshot.}
\end{figure}
\subsection{Limitations and Future work}

In the current form, the proposed system has a couple of limitations. First, we only have 2D control over the HIFU focal point and a limited focal distance due to the limited space in our small animal scanner. This limitation could be overcome using 3D HIFU actuation systems with larger and variable focal distance in clinical MRI scanners\cite{dai2021robotic}. The second limitation of the system demonstrated in this work is that while we can localize the capsule using the magnetic artifact, we cannot visualize the drug release in MR images, like \cite{matoori2019mri} and the drug release was controlled in a open-loop manner by estimating through release rate. Due to the relatively large size of our magnetic core, the magnetic image artifact seen in Fig. 5 is as large as the field of view of the camera in Fig. 6. Therefore, the artifact obstructs the visual drug release feedback in MR images. In order to mitigate the effect of artifact, the magnetic core size can be decreased down to \SI{300}{\micro\meter} by using a smaller capsule \cite{tiryaki2021magnetic}. Downscaling the capsule size would also ease the navigation of the capsule in the narrow region of the smaller intestine.

However, downscaling requires further investigation physical model of the bubble removal and hydrodynamics of drug release in future work. While we experimentally demonstrated the effect of the hole geometry on required pressure, an accurate dynamic model of the bubble removal process is needed for scaling down the air bubble-based stopper mechanism. Similarly, the correlation between acoustic pressure and release rate is evident in our experiments; however, our results do not explain the exact mechanism for drug release. 

Besides modeling, downscaling also brings other practical issues, such as drug loading and balancing acoustic radiation forces. Currently, we use \SI{300}{\micro\meter}-diameter commercial needles to load the drug. However, we would need custume-made needles to load drugs through the smaller hole with a more precise injection mechanism. Similarly stable magnetic actuation would be more challenging in smaller scales  due to the acoustic radiation forces. In this study, we used relatively large scaled capsule robot with high magnetic force to acoustic radiation force ratio. However, down-scaled capsule would be more affected by radiation force and prevent release. The affect of radiation force could be counteracted by the closed-loop magnetic actuation with high frequency imaging actuation sequences \cite{tiryaki2021magnetic,tiryaki2020realistic}.

Another important future work is to characterize the drug release efficiency in more realistic environments, such as physiologically relevant fluids and tissue-like acoustic environments. In this study, we only investigated in vitro conditions with non-biological fluids in room temperature, like \cite{jeong2020acoustic}  We expect fluid properties like viscosity and pH to affect the drug release rate. Therefore, a more comprehensive study with body fluids is necessary. In addition, inhomogeneous acoustic transmission through different tissues and moving organs would affect acoustic power transfer efficiency. In future, MR thermometery-based feedback can be used calculate the exact position and pressure of the HIFU focus point \cite{dai2021robotic}.

\section{Acknowledgement}
The authors thank U. Bozüyük, M. Han, S.O. Demir, A. Aghakhani, and A. Bhargava for their theoretical discussions.

\balance
\bibliography{IEEEabrv,reference.bib}
\bibliographystyle{IEEEtran}

\end{document}